\documentclass{ecai}
\paperid{1406}

\usepackage{latexsym}
\usepackage{amssymb}
\usepackage{amsmath}
\usepackage{amsthm}
\usepackage{booktabs}
\usepackage{enumitem}
\usepackage{graphicx}
\usepackage{color}

\theoremstyle{plain}
\newtheorem{theorem}{Theorem}[section]

\theoremstyle{definition}

\theoremstyle{remark}
\newtheorem{remark}[theorem]{Remark}

\newcommand{\BibTeX}{B\kern-.05em{\sc i\kern-.025em b}\kern-.08em\TeX}

\usepackage{microtype}

\usepackage{hyperref}
\usepackage{url}
\usepackage{subcaption}

\usepackage{mathtools}

\usepackage{xcolor}
\usepackage{textcomp}
\usepackage{tikz}

\newcommand{\redcircle}{\tikz\draw[red,fill=red] (0,0) circle (0.5ex);}
\newcommand{\greencircle}{\tikz\draw[green,fill=green] (0,0) circle (0.5ex);}

\newcommand{\rX}{{\color{red}\texttimes}}

\begin{document}

\begin{frontmatter}

\title{Forward Only Learning for Orthogonal Neural Networks of any Depth}


\author[A]{\fnms{Paul}~\snm{Caillon}\thanks{Corresponding Author. Email: paul.caillon@dauphine.psl.eu}\footnote{Equal contribution}}
\author[A]{\fnms{Alex}~\snm{Colagrande}\footnotemark[1]}
\author[A]{\fnms{Erwan}~\snm{Fagnou}} 
\author[A]{\fnms{Blaise}~\snm{Delattre}} 
\author[A,B]{\fnms{Alexandre}~\snm{Allauzen}} 
\address[A]{Miles Team, LAMSADE, Université Paris-Dauphine - PSL, Paris, France}
\address[B]{ESPCI PSL, Paris, France}
\begin{abstract}
    Backpropagation is still the de facto algorithm used today to
    train neural networks.
    With the exponential growth of recent architectures, the
    computational cost of this algorithm also becomes a burden. The
    recent PEPITA and forward-only frameworks have proposed promising
    alternatives, but they failed to scale up to a handful of hidden
    layers, yet limiting their use.
    In this paper, we first analyze theoretically the main limitations of
    these approaches. It allows us the design of a forward-only
    algorithm, which is equivalent to backpropagation under the linear
    and orthogonal assumptions. By relaxing the linear assumption, we
    then introduce FOTON (Forward-Only Training of Orthogonal Networks)
    that bridges the gap with the backpropagation
    algorithm. Experimental results show that it outperforms PEPITA,
    enabling us to train neural networks of any depth, without the need
    for a backward pass.  
    Moreover its performance on convolutional networks clearly opens up avenues for its application to more
    advanced architectures. The code is open-sourced on \href{https://github.com/p0lcAi/FOTON}{github}. 
\end{abstract}
 \end{frontmatter}


\section{Introduction and Related Work}

Backpropagation (BP)~\citep{backprop} is still the
cornerstone for training deep neural networks.  With the exponential
growth of architectures, its limitations have been highlighted, notably
regarding its lack of efficiency, along with its biological
implausibility \citep{lillicrap2020backpropagation}. More
specifically, the sequentiality of the backward pass represents a
computational bottleneck when training deep networks. 
The memory footprint and execution time represent a significant
drawback and reducing this burden is a major challenge that could
allow the use of state-of-the-art models on resource-limited devices
\citep{khacef2023spike, kendall2020training}.

Alternatives have thus been proposed, most targeting the backward phase. Feedback Alignment (FA) replaces exact gradients with random feedback directions \citep{lillicrap2016random}, while Direct FA (DFA) transmits feedback directly from the output to each layer in parallel \citep{nokland2016direct}. These simplifications reduce test accuracy, though later works attempt to narrow the gap with BP by partially mimicking its behavior \citep{akrout2019deep, xiao2018biologically, lansdell2019learning, Guerguiev2020Spikebased, kunin2020two}. Nonetheless, clear limitations remain: poor scaling to deep convolutional networks \citep{bartunov2018assessing, launay2019principled, moskovitz2018comparison, crafton2019direct}, and the need for an exact backward pass in certain modules, such as attention layers in transformers \citep{launay2020direct}.

More recently, forward-only (FO) algorithms have been proposed as an
alternative to the BP, replacing the backward pass by a modulated second forward pass.
A first line of work relies on the estimation of the gradient, computed locally from a modified
forward pass using directional derivatives. These forward gradients provide a plausible update for the parameters 
\citep{baydin2022gradients,silver2021learning,margossian2019review,fournier2023can}.
For efficient computation, the Jacobian-vector product is involved during the forward
pass. 
However, while forward gradients are unbiased, they exhibit a large
variance. This inhibits the convergence of the training as well as the
scalability of this approach even in combination with local losses
\citep{ren2022scaling,belouze2022optimization,fournier2023can}.

In this work, we focus on another family of forward-only algorithms, in which a second pass or \emph{modulated forward pass}~\citep{srinivasan2023forward} computes network outputs on a perturbed input. The nature of this perturbation varies by method: for example, \citet{hinton2022forward} use “negative” data samples, requiring designing or collecting a special dataset, whereas the PEPITA framework \citep{dellaferrera2022error} derives its modulation directly from the error signal, allowing a more agnostic and data-efficient approach.  

\citet{srinivasan2023forward} further analyzed the learning dynamics of these forward-only schemes and extended PEPITA to slightly deeper architectures but still found that, in practice, they fail to scale beyond five hidden layers and cannot adapt to convolutions. As a result, existing methods remain limited to shallow, fully connected networks.  

To address these limitations, we identify the core bottleneck: forward-only passes cannot reliably transmit gradient information through many layers, and the absence of a true backward path demands special data or ad-hoc fixes. We show that under linear orthogonality, a purely forward strategy exactly recovers BP updates. Leveraging this, we introduce \emph{FOTON}, which matches BP in the orthogonal linear regime and scales robustly to deep non-linear networks. 
Crucially, FOTON dispenses with any backward pass or storage of the automatic differentiation graph: all updates are computed via forward evaluations alone, dramatically reducing memory overhead by \textbf{eliminating the need to retain activations or gradient buffers}.  
Our  contributions are:
\begin{itemize}[noitemsep,topsep=0pt,parsep=0pt,partopsep=0pt]
  \item We show theoretically that, for orthogonal linear networks, PEPITA’s update rules can be modified to exactly reproduce BP’s gradients.
  \item We relax the linearity constraint to derive FOTON, a forward-only algorithm that continues to deliver effective learning signals in deep non-linear models without ever invoking a backward pass.
  \item We provide an in-depth empirical study of FOTON’s training dynamics, illustrating its stability and scalability compared to PEPITA.
  \item We adapt FOTON to convolutional architectures—orthogonalizing conv kernels via BCOP—and demonstrate, for the first time, successful forward-only training of multi-layer convnets.
\end{itemize}

\begin{figure*}
		\centering
		\includegraphics[width=\textwidth]{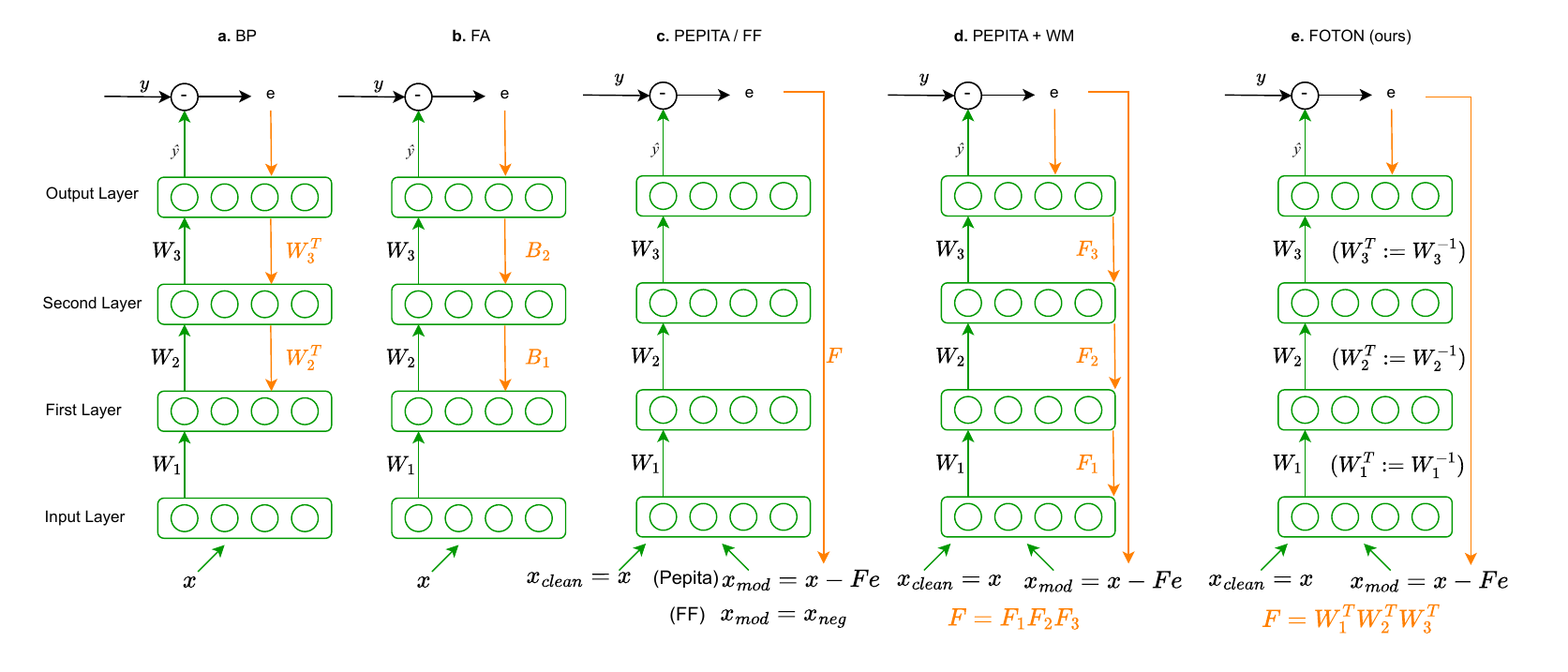}
		\caption{Overview of different error transportation configurations, inspired by \citep{dellaferrera2022error}. 
			a) Backpropagation (BP). b) Feedback Alignment (FA). c) Forward Only propositions (PEPITA and Forward-Forward).
         d) Adapted PEPITA (PEPITA+Weight Mirroring), from \citep{srinivasan2023forward}. e) Forward-Only Training of Orthogonal Networks (FOTON). 
			Green arrows indicate forward paths, orange ones error paths. Weights are denoted $W_\ell$, feedback matrices are
			$B_\ell$ if specific to a layer (FA). 
         Matrix used to project the error on the input layer are 
         denoted as $F_\ell$ if specific to a layer (PEPITA+WM from \citep{srinivasan2023forward}) or $F$ if global (standard PEPITA). In the case of FOTON
         the $F_\ell$ are replaced by $W_\ell^\top:=W_\ell^{-1}$ as the layers are orthogonal.}
		\label{fig:sum}
	\end{figure*}
    
\section{Background and Notations}
\label{sec:related}
	
Let $f(x; \theta)$ be a neural network with $L$ layers, where
$x \equiv h_0$ is the input and $\theta$ denotes its parameters.  The
network computes the prediction $\hat{y}=a_L$,
where $a_\ell = W_\ell h_{\ell-1}$ and $h_\ell = \sigma_\ell(a_\ell)$
are respectively  pre-activation and activation of the layer $l$ (for $ 1 \leq l \leq L$), and 
$\sigma_\ell$ denotes the element-wise non-linear activation function.  Given a loss
function $\mathcal{L}$,
$\nabla_\ell \mathcal{L} \equiv \frac{\partial \mathcal{L}}{\partial
  h_\ell}$ represents the gradient of the loss with respect to the
activation at layer $\ell$.

The BP algorithm computes this term sequentially after a forward
pass, starting from the output layer. The term
$e \equiv \nabla_L \mathcal{L}$ corresponds to the error made by the
network for the input $x$. For the Mean Squared Error (MSE) loss, it can be
written as $e = \hat{y} - y$. Given a learning rate $\eta$ and
denoting the Hadamard product by $\odot$, the weight updates are
written as follows:
\begin{align}
   \delta W_\ell &= -\eta \Bigl(\delta h_\ell^{\mathrm{BP}} \odot 
  \sigma_\ell^\prime(a_\ell)\Bigr)
h_{\ell-1}^\top, \ \forall \ell < L, \label{eq:bp_in} \\
  \delta W_L &= - \eta \: e \: h_{L-1}^\top,  \label{eq:bp_out} \\
  \quad \delta h_\ell^{\mathrm{BP}} &= \nabla _\ell \mathcal{L} = W_{\ell+1}^\top
\Bigl(
  \delta h_{\ell+1}^{\mathrm{BP}}
\odot
  \sigma_{\ell+1}^\prime(a_{\ell+1})
\Bigr), \forall \ell.          
\end{align}
    
Hence, the update of a layer $\ell$ depends on the error back-propagated through the $\{L, L-1, ..., \ell+1\}$ subsequent layers, as illustrated in Figure~\ref{fig:sum}.
It also requires the storage of the computation graph to estimate the gradients in reverse order. This overhead may becomes prohibitive  for deep networks.
It is also worth noting that the weight updates are ``backward-locked'', meaning that it 
is not possible to update the weights of a layer before the gradients of the subsequent layers have been computed.

In the PEPITA framework introduced by \citet{dellaferrera2022error}, the error signal is projected in the input space to compute a second
\textit{modulated} forward pass.
In the following, we adopt the formalism of \citet{srinivasan2023forward}.
While the first forward pass is applied to the input $x$, the second forward pass uses $x-Fe$ as the input:
\begin{equation}\label{eq:modulatedpass}
    \begin{split}
   &h_1^{err} = \sigma_1(a_1^{err}) = \sigma_1 (W_1 (x - F e) ), \\
    &h_\ell^{err} = \sigma_\ell(a_\ell^{err}) = \sigma_\ell (W_\ell h_{\ell-1}^{err})  \; \text{ for } 2\leq \ell \leq L,
    \end{split}
\end{equation} 
where $F$ is a fixed random matrix.
After two forward passes, the weights are updated using the PEPITA learning rule:
\begin{align}
   \delta W_1 & = -\eta \: (h_1-h_1^{err}) (x - Fe)^\top; \label{eq:pepita_dw_input}\\
   \delta W_\ell & = - \eta \: (h_\ell-h_\ell^{err}) (h_{\ell-1}^{err})^\top \;
   \text{ for } 2\leq \ell< L; \label{eq:pepita_dw_hidden} \\
   \delta W_L & = - \eta \: e \: (h_{L-1}^{err})^\top. \label{eq:pepita_dw_out}
   \end{align}

\begin{table*}[ht!]
   \caption{Test accuracy [\%] achieved by BP, BP with orthogonalized weights as each step (denoted BP $\perp$), FA, DFA, PEPITA, PEPITA + WM and FOTON on the MNIST dataset on 
   fully connected networks of different depths.
   Used hyperparameters are reported in the Appendix~\citep{supplementary_mat}. Bold fonts refer to the best results for alternatives to BP.
   \rX~represents algorithms that failed to scale to the corresponding depth.}
   \centering
   \begin{tabular}{lccccccc}
       \toprule
       \textbf{Learning rule} & \textbf{Forward-only} & \textbf{1 HL} & \textbf{2 HL} & \textbf{3 HL} & \textbf{5 HL} & \textbf{10 HL}& \textbf{50 HL} \\
       \midrule
       BP via SGD      &  \redcircle  & 98.91 & 98.85 & 98.89 & 98.49 & 97.44 & 97.49\\
       BP $\perp$     &  \redcircle& 94.33 	&95.19 	&96.18 	&96.91 	&96.5 & 97.85\\
       \midrule
       FA              & \redcircle & 98.42 & 98.25 & 98.08 & 97.96 & 95.43 &  \rX\\
       DFA             & \redcircle & 98.32 & 98.26 & 98.30 & 98.10 & 94.50 & \rX\\
       PEPITA (original)        & \greencircle & 98.01 & \rX & \rX & \rX & \rX & \rX\\
       PEPITA  + WM    & \greencircle & 98.42 & 98.19 & 96.33 & \rX & \rX & \rX\\
       FOTON (ours)    & \greencircle & \textbf{98.49} & \textbf{98.61} & \textbf{98.36} & \textbf{98.32} &  \textbf{97.27} & \textbf{88.42} \\
       \bottomrule
   \end{tabular} 
   \label{tab:MNIST}
\end{table*}

\section{Forward-Only Training of Orthogonal Networks (FOTON)}
\label{sec:foton}        

Our contribution relies on the key insight that, in orthogonal linear networks, the PEPITA update rules (Eqs.~\ref{eq:pepita_dw_input}–\ref{eq:pepita_dw_out}) can be made to reproduce exactly the gradients of standard backpropagation (Eqs.~\ref{eq:bp_in},\ref{eq:bp_out}). By relaxing the linearity assumption, we derive FOTON, a forward‐only training algorithm that preserves this equivalence in the linear regime while still delivering strong learning signals in deep non‐linear networks, as demonstrated by our experiments on MNIST, CIFAR‐10 and CIFAR‐100 (see Section~\ref{sec:results}).  

Although naively recomputing the error‐projection matrix \(F\) at every iteration might add a pseudo extra forward pass—so FOTON is not inherently faster per step than backpropagation—its real benefit is the elimination of the backward pass and the entire computation graph. This pure forward execution makes FOTON especially attractive in memory‐constrained settings, where removing the backward lock and all intermediate activation storage brings substantial practical gains. Moreover, as shown in Section~\ref{sec:ablation}, one can further reduce this overhead by relaxing the exact weight‐alignment requirement by updating \(F\) less frequently or via weight‐mirroring schemes, aligning forward and feedback weights, as in~\citep{akrout2019deep}.

\subsection{Limitations of the PEPITA learning rule}

Introduced by \citet{dellaferrera2022error}, the PEPITA learning rule illustrated in Figure~\ref{fig:sum}c is a promising alternative to BP.
Initially tested on only two-layer networks, it has been extended 
to networks with 3 hidden layers and even
5 hidden layers in very specific settings by 
\citet{srinivasan2023forward} as illustrated in Figure~\ref{fig:sum}d.
However, the empirical results show a trend of decreasing (and even collapsing) performance with the depth of the network, contrary
to what is observed with BP. Furthermore, the theoretical analysis
provided in \citep{srinivasan2023forward} is only valid for two-layer networks.
For deeper networks,  commonly used in practice, 
a theoretical analysis of the convergence and learning dynamics 
is made near impossible due to the progressive degradation of the signal
during the modulated forward. 
The results and analysis provided in Section~\ref{sec:results} support the observation that the modulated forward pass is unable to transmit a meaningful error signal.

\subsection{FOTON learning rule}
First, instead of writing $e = \hat{y} - y$ as in the previous works
on PEPITA, we use the more general form $e = \nabla_L
\mathcal{L}$. This formulation allows us to consider any differentiable loss, hence generalizing the work of \citet{dellaferrera2022error,srinivasan2023forward}. We then use the following changes in the PEPITA update rules~\ref{eq:pepita_dw_input}–
\ref{eq:pepita_dw_out}, as in \citep{srinivasan2023forward}:
\begin{align}
   \delta W_\ell & = - \eta \: (h_\ell-h_\ell^{err}) (h_{\ell-1})^\top \;
   \text{ for } 1\leq \ell< L; \label{eq:foton_in} \\
   \delta W_L & = - \eta \: e \: h_{L-1}^\top \: . \label{eq:foton_out}
   \end{align}
   With these modifications, the update rule closely relates to the
   one calculated by BP, and in particular matches BP for the last layer (equations~\ref{eq:foton_out} and
   ~\ref{eq:bp_out}). Furthermore, if
   $h_\ell-h_\ell^{err} = \nabla_\ell \mathcal{L}\text{ for } 1\leq
   \ell< L$, the update given by equation~\ref{eq:foton_in} resolves
   exactly to the one given by equation~\ref{eq:bp_in}.  Therefore to
   reduce the gap with BP, the quantity 
   $\delta h_\ell ^{FO}:=h_\ell-h_\ell^{err}$ becomes pivotal. Let us detail its computation: 
\begin{align}
   \delta h_\ell ^{FO}=& \sigma_l(W_\ell\sigma_{\ell-1}(W_{\ell-1}\sigma_{\ell-2}( \cdots \sigma_1(W_1x))))    \label{eq:deltafoton}
\\
   -& \sigma_l(W_\ell\sigma_{\ell-1}(W_{\ell-1}\sigma_{\ell-2}( \cdots \sigma_1(W_1(x-Fe))))) \nonumber
\end{align}
In the next section, we will consider different assumptions to improve PEPITA.

\subsubsection{Orthogonal linear networks}
In the case of linear networks, equation~\ref{eq:deltafoton} reads:
\begin{equation}
\delta h_\ell ^{FO}= W_\ell W_{\ell-1}\cdots W_1 Fe \label{eq:fotonortho}
\end{equation}
Assuming a perfect weight alignment in the sense proposed by \citet{srinivasan2023forward} following the 
idea of \citet{akrout2019deep} (\textit{i.e.} $F=W_1^\top W_2^\top \cdots W_L^\top$), and orthogonal weights (\textit{i.e.} $W_\ell^\top W_\ell=I$), 
equation~\ref{eq:fotonortho} simplifies to:
\begin{equation}
   \delta h_\ell ^{FO}=W_{\ell+1}^\top \cdots W_{L}^\top e = \delta h_\ell ^{BP}.
   \label{eq:fotonorthofinal}
\end{equation}  
This means that in the case of orthogonal linear networks of any depth, assuming a perfect weight alignment of F
with the forward weights, FOTON is strictly equivalent to BP.

\subsubsection{Non-linear networks}



In the case of non-linear networks, we show empirically in section~\ref{sec:results} that the FOTON learning rule remains very effective. While there is no theoretical guarantee of convergence in this general case, we can still get some intuition as to why these results are observed.

By approximating the activation functions linearly ($\sigma'_\ell(x) \approx \lambda$), we can see how FOTON approaches backpropagation up to a rescaling:
\begin{equation}
       \delta h_\ell ^{FO} \approx \lambda^\ell W_{\ell+1}^\top W_{\ell+2}^\top \cdots W_L^\top e \approx \lambda^{2\ell-L+1} \cdot \delta h_\ell ^{BP}
       \label{eq:fotonnonlin}
\end{equation}
This shows the direction provided by FOTON is still a descent direction, as showed in Figure~\ref{fig:cos_sim_nonlinear}, where the signal given by FOTON has a positive cosine similarity with the true gradient. The constant factor can be absorbed by the learning rate or an adaptive optimizer like Adam. See the Appendix~\citep{supplementary_mat} for more details.

\subsubsection{Convolutional networks}
In the case of convolutional layers, the update rule must be adapted
since the outer product of equation~\ref{eq:foton_in} gives a valid
update for a dense layer but not for the convolutional kernel.  To the
best of our knowledge, the only method proposed in the literature that
handle convolution layers in the forward-only framework is introduced
in~\cite{dellaferrera2022error}, where they average the products
between the output components and the respective convolutional
receptive fields. However, this method fails to scale to deep networks. In this work we propose a  different approach that
leverages the strong analogy between BP and FOTON, with the rule to update  a convolutional kernel: 
\begin{align*}
   \delta k_{\ell} = - \eta \text{conv}_{h_{l-1}}^{\star} (h_l - h_l^{\text{err}})
\end{align*}
where $\text{conv}_k(x) = k * x$ is the left-convolution by a kernel
$k$ and $\star$ indicates the adjoint operator (\textit{i.e.} the transposed convolution).
We highlight that our proposition drastically differs from the
previously cited method, since it allows us to still match
backpropagation in the linear case, and to remain first order
approximation in the general case. To ensure the orthogonality of the
convolutional layers, we use Block
Convolution Orthogonal Parametrization (BCOP) as introduced by
\citet{li2019preventing}. Lastly, we only consider average pooling
layers as the max pooling operation is not adapted to the forward-only
framework. More details are provided in the Appendix~\citep{supplementary_mat}.

\begin{table*}[ht!]
   \caption{Test accuracy [\%] achieved by BP, PEPITA and FOTON on the CIFAR-10/100 datasets on fully connected networks of different depths, over 5 independent runs.
   The results reported for PEPITA are the best results obtained following the improvements of PEPITA done by \citet{srinivasan2023forward}.
   \rX~represents algorithms that failed to scale to the corresponding depth.}
   \centering
   \resizebox{\textwidth}{!}{%
   \begin{tabular}{ccccccc|cccccc}
      \toprule
      & \multicolumn{6}{c|}{\textbf{CIFAR-10}} & \multicolumn{5}{c}{\textbf{CIFAR-100}} \\
      \textbf{Learning rule} & \textbf{1 HL} & \textbf{2 HL} & \textbf{3 HL} & \textbf{5 HL} & \textbf{10 HL} & \textbf{50 HL}& \textbf{1 HL} & \textbf{2 HL} & \textbf{3 HL} & \textbf{5 HL} & \textbf{10 HL} & \textbf{50 HL}\\     
      \midrule
      BP via SGD & 56.23 & 57.42 & 57.51 & 52.55 & 52.96 & 45.62&  29.36 & 29.80 & 29.01 & 28.32 & 25.15 & 13.53  \\
      BP $\perp$ & 55.78 & 54.90 & 54.68& 52.97 & 52.53 & 52.05&  28.15 & 28.30 & 28.02 & 28.20 & 25.49 & 20.88 \\
      \midrule
      PEPITA (best results) & 53.80 & 53.44 & 53.80 & \rX & \rX & \rX & \textbf{27.07} & \textbf{26.95} & 23.13 & \rX & \rX & \rX \\
      FOTON (ours) & \textbf{55.61} & \textbf{55.14} & \textbf{55.70} & \textbf{53.09} & \textbf{52.22} & \textbf{40.86}&25.23 & 25.02 & \textbf{23.84} & \textbf{22.60} & \textbf{21.00}&\textbf{15.65} \\
      \bottomrule
  \end{tabular}}
  \label{tab:CIFAR}
\end{table*}
\subsection{Layer orthogonalization}
FOTON ensures left-orthogonality of the \textbf{weight matrices} throughout
training via the Björck algorithm \citep{bjorck} as done by \citet{li2019preventing}. However, our
method does \textit{not} depend on this peculiar choice and any orthogonalization scheme—e.g.\ Cayley transforms \citep{trockman2021orthogonalizing} or skew‐symmetric exponentials \citep{singla2021skew}—could be substituted.
This orthogonalization is carried out at each iteration, ensuring near-perfect orthogonality throughout training. While it introduces a non-trivial runtime overhead, substantial benefits arise from enforcing orthogonality. 

First, orthogonal layers preserve the norm of backpropagated gradients, preventing both vanishing and exploding gradients and stabilizing activation distributions \citep{pmlr-v97-anil19a,2016arXiv161101967R}. 
Second, maintaining unit singular values yields well‐conditioned optimization landscapes that can accelerate convergence and improve generalization \citep{saxe2013exact}.

Third, enforcing orthogonality provides provable robustness guarantees: by bounding the Lipschitz constant, networks become more resistant to adversarial perturbations and admit deterministic certificates of robustness \citep{tsuzuku2018lipschitz}. 
Finally, orthogonal transformations preserve pairwise distances in feature space, promoting stable feature reuse and mitigating representation collapse in very deep architectures as empirically verified in the deepest models used in Tables\ref{tab:MNIST},\ref{tab:CIFAR}. 

We emphasize that the left-orthogonality constraint applies to rectangular matrices, ensuring that column vectors remain orthogonal throughout training. Despite the added cost, this structural regularization allows deeper or more compact models to train reliably.  We investigate a potential reduction of this overhead by updating the orthogonalization less frequently  in our ablation study Section~\ref{sec:ablation}.

\section{Empirical Evaluation}
\label{sec:results}

\subsection{Experimental Setup}
In this section, we detail our empirical evaluation of FOTON and present its results.
All experiments were run on a single NVIDIA A100 GPU, and all results report the mean (with negligible standard deviation) over five independent runs. We evaluate on MNIST, CIFAR-10/100 using standard train/test splits without any data augmentation.

For our multilayer perceptrons, we consider three regimes: shallow networks with 1–3 hidden layers of 1 024 units and ReLU activations (following \citet{srinivasan2023forward}); mid-depth models with 5 or 10 layers of 256 units and ReLU; and very deep stacks of 50 layers with 256 units and Tanh activations (to avoid the “dying ReLU” issue in extreme depth).

In convolutional settings, we use two successive 3×3 conv layers (32 then 64 filters), each followed by average pooling, capped by a single fully-connected classification layer; all convolutional kernels are orthogonalized via BCOP.
Across all architectures, we compare FOTON to PEPITA (and its variants) and to standard BP (with or without per-step orthogonalization). The complete set of optimization hyperparameters is provided in the Appendix~\citep{supplementary_mat}. Finally, we supplement these results with the analysis of the training dynamics of FOTON when compared to PEPITA and ablation studies on the frequency of layer orthogonalization and on how often the $F$ matrix is recomputed.

\subsection{Results}
\label{subsec:results}

We begin by evaluating FOTON on non‐linear networks of varying depth—both fully connected and convolutional—across the MNIST, CIFAR-10, and CIFAR-100 benchmarks, and compare its accuracy to that of PEPITA and standard backpropagation. Our experimental protocol closely follows \citep{srinivasan2023forward}, using the same tasks, hyperparameters, and shallow‐network configurations as slight deviations in those parameters caused PEPITA's training to collapse. Crucially, despite extensive tuning and the enhancements from \citet{srinivasan2023forward}, we were unable to train PEPITA beyond five hidden layers, whereas FOTON scales reliably to much deeper architectures.

\subsubsection{Results on Fully Connected Networks}
The results regarding fully connected networks are summarized in Table~\ref{tab:MNIST} for MNIST and Table~\ref{tab:CIFAR} for CIFAR-10 and CIFAR-100. First, we observe and experimentally confirm that FOTON provides learning to deep non-linear networks, answering to a 
major limitation of PEPITA. This confirms the theoretical analysis made in Section~\ref{sec:foton}.
Our results demonstrate that FOTON scales reliably to depths of up to 50 layers, whereas PEPITA fails beyond five layers despite exhaustive hyperparameter tuning. On MNIST (Table~\ref{tab:MNIST}), FOTON achieves 98.32\% at 5 hidden layers, 97.27\% at 10 layers and still 88.42\% at 50 layers, compared with PEPITA’s collapse after 3 layers (96.33\% at 3 HL, no result beyond). 

Similarly, on CIFAR-10 (Table~\ref{tab:CIFAR}), FOTON reaches 55.70\% at 3 layers, 53.09\% at 5 layers and 52.22\% at 10 layers, while PEPITA cannot be trained past 3 layers. In convolutional settings (Table~\ref{tab:CNNs}), FOTON attains 90.69\% on a 2×conv MNIST network and 28.48\% on a 1×conv CIFAR-100 network, outperforming PEPITA in every configuration. Moreover, on fully-connected models up to 10 layers, FOTON matches or exceeds standard backpropagation (and even approaches orthogonal-BP performance) without ever executing a backward pass and even more crucially \textbf{without needing to store any computation graph}.

Finally, these experiments omit data augmentation, batch normalization and advanced learning-rate schedules or optimizers as Adam~\citep{kingma2014adam}—techniques that typically improve backpropagation—so integrating them represents a clear path to further boost FOTON’s accuracy.  

\begin{figure*}[ht]
    \centering
        \includegraphics[width=\linewidth]{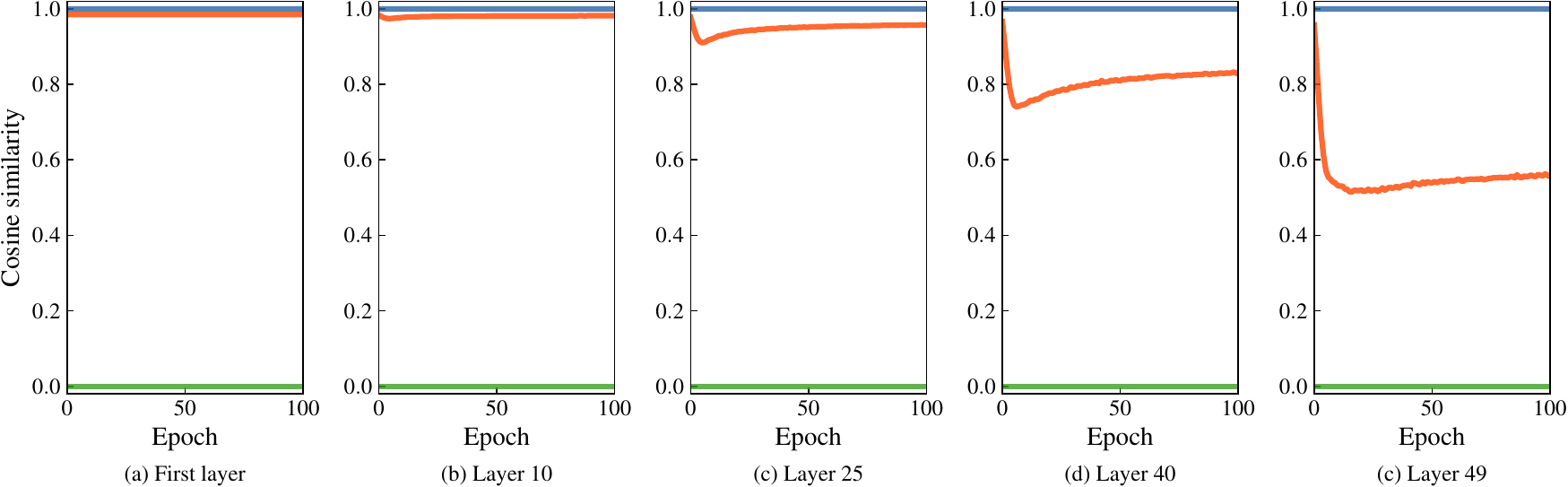}
        (a) Case of stable training.

        \vspace{0.4cm}
        
        \includegraphics[width=\linewidth]{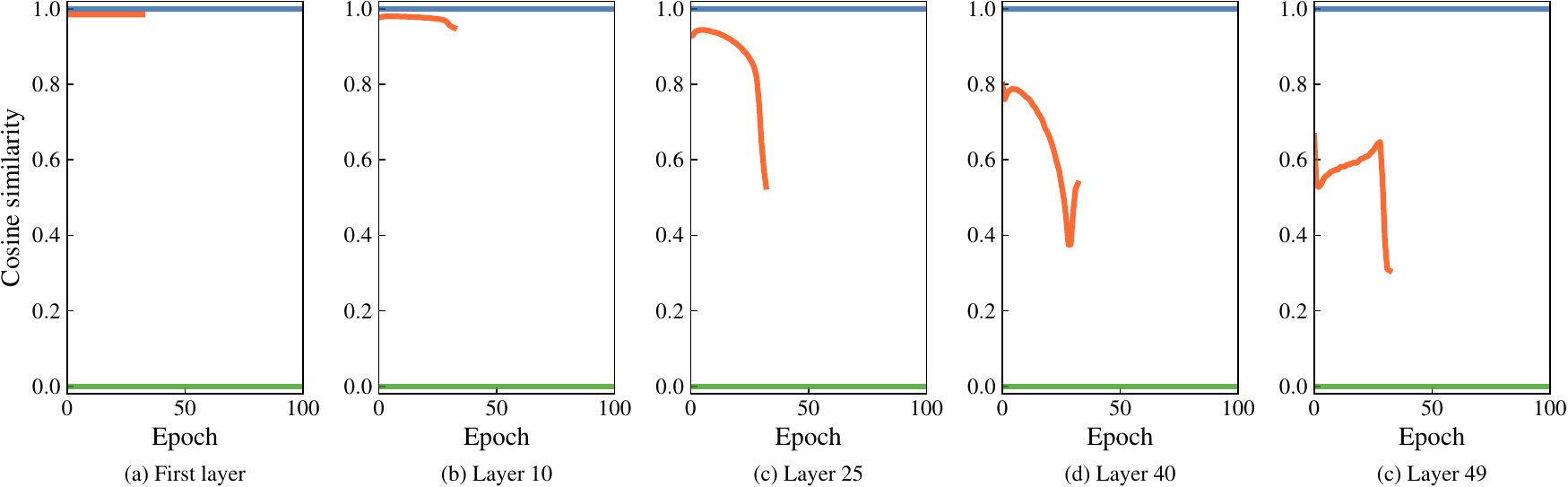}
        (b) Case of unstable training.
    
    \caption{Cosine similarity between the true gradient computed by BP and the estimates computed using FOTON (blue), 
   PEPITA (green), and PEPITA with orthogonal initialization (orange), on a 50 hidden layers linear network on MNIST
   during training.
   The cosine similarity is averaged over the training set, and is reported for 5 layers across the network.
   Two cases are reported depending on the stability of PEPITA training (stable in~(a) and unstable in~(b)).
   Stable training was obtained for a very small learning rate $lr=1e^{-5}$.}
   \label{fig:cos_sim_linear}
\end{figure*}

\subsubsection{Results on Convolutional Networks}
Extending FOTON to convolutional architectures requires reconciling the orthogonality constraint—which is naturally formulated on the large Toeplitz matrix of the convolution with the small, spatial structure of the kernel itself. 
To enforce orthogonality on each convolutional layer we employ Block Convolution Orthogonal Parametrization (BCOP) \citep{li2019preventing}. 
Unlike the original PEPITA update, which correlates post and pre-synaptic activations patch-by-patch and does not align with the true gradient of the linearized conv layer, FOTON’s convolutional update is derived by mimicking backpropagation: in the linear (or piecewise-linear) regime, the exact weight gradient is given by applying the adjoint convolution (i.e.\ the transposed convolution) of the error to the input feature map.

In practice, we implement this via PyTorch’s \texttt{torch.nn.grad.conv2d\_input} operator. 
This approach recovers the true backprop gradients under the orthogonality assumption and enables training multi-layer convolutional networks entirely in a forward-only fashion. Further implementation details are provided in the supplementary material.

\begin{table}[ht!]
   \caption{Test accuracy [\%] achieved by BP, PEPITA and FOTON on the MNIST and CIFAR-100 datasets, on convolutional networks with one or two convolutional layers, and a linear classifier.
   \rX~represents algorithms that failed to scale to the corresponding depth.}
   \centering
   \resizebox{\columnwidth}{!}{%
   \begin{tabular}{c|c|c|c|cccc}
      \toprule
      & \multicolumn{2}{c|}{\textbf{Conv $\times 1$}} & \multicolumn{2}{c}{\textbf{Conv $\times 2$}} \\
      \textbf{Learning rule} & \textbf{MNIST} & \textbf{CIFAR-100} & \textbf{MNIST}  & \textbf{CIFAR-100} \\     
      \midrule
      BP via SGD  & 98.86 &  34.20 &  98.93 & 34.45\\

      PEPITA (best results)  & \textbf{98.29} & 27.56 & \rX  & \rX \\
      FOTON (ours) & \textbf{98.29} & \textbf{28.48} & \textbf{90.69} &  \textbf{21.98} \\
      \bottomrule
  \end{tabular}}
  \label{tab:CNNs}
\end{table}

Table~\ref{tab:CNNs} summarizes FOTON’s performance on convolutional architectures with one or two convolutional layers for MNIST and CIFAR-100. To our knowledge, this is the first demonstration of a forward-only method successfully training networks with more than a single convolutional layer. Although FOTON does not yet match standard backpropagation’s accuracy, it clearly outperforms PEPITA and scales to deeper conv nets, a trend consistent with its superior behavior in fully connected settings, even for shallow models, showcasing
the potential of the algorithm for more advanced architectures.

 \subsection{Analysis of the training dynamics}
 \label{sec:trainingdynamics}
 In this section we conduct an analysis of the difference of the training dynamics between FOTON and PEPITA, in terms of the updates computed.
 We begin with deep linear networks—where FOTON is theoretically equivalent to backpropagation—to validate our analysis, and then extend the study to non-linear models to shed light on the behaviors underlying the results reported in Section~\ref{subsec:results}.
 
 \begin{figure*}[ht]
   \centering
   \includegraphics[width=\linewidth]{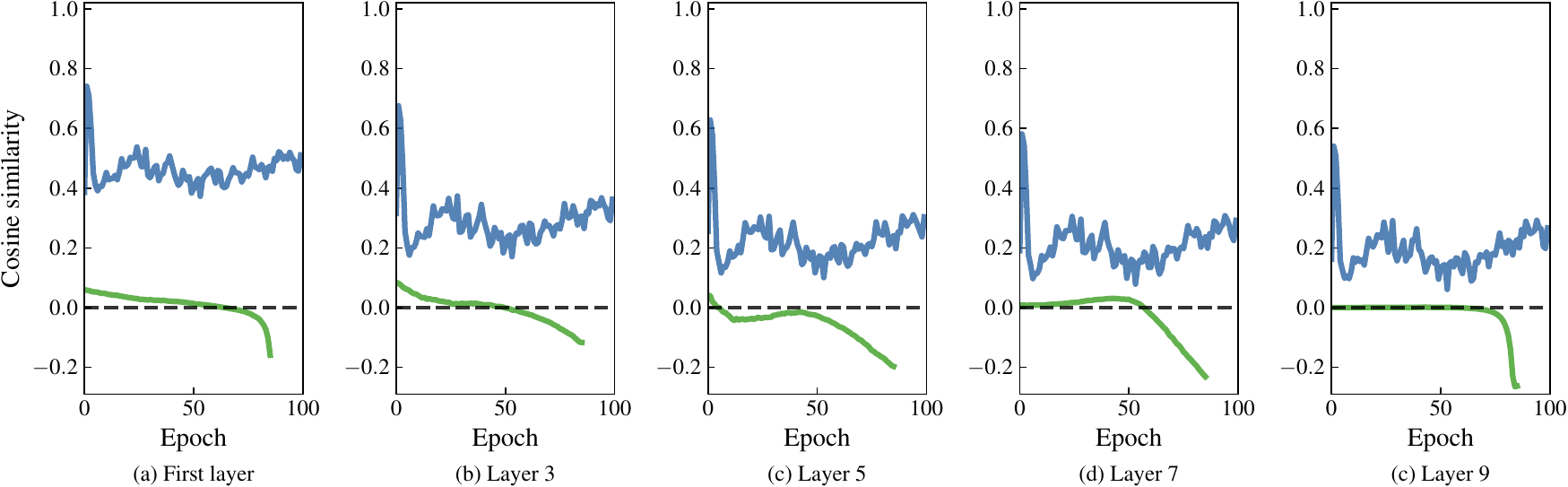}
   \caption{Cosine similarity between the true gradient computed by BP and the estimates computed using FOTON (blue) and
   PEPITA (green), on a 10 hidden layers network with ReLU nonlinearity on MNIST
   during training.
   The cosine similarity is averaged over the training set, and is reported for 5 layers across the network.}
   \label{fig:cos_sim_nonlinear}
\end{figure*}
 \subsubsection{Linear case}
 As shown in Figure~\ref{fig:cos_sim_linear}, in the linear case the cosine similarity between the gradients computed by FOTON and BP is exactly 1
 for all layers, meaning that FOTON computes the same gradients as BP. This is coherent with the theoretical analysis made in Section~\ref{sec:foton}, 
 as in this case, the update rules of BP and FOTON are exactly the same.
 When comparing the cosine similarity between the gradients computed by BP, and PEPITA initialized following 
 the scheme developed in \citep{srinivasan2023forward}, we observe that it is very close to 0, meaning that PEPITA
 gives directions that are close to orthogonal to the true gradient.
 This suggests that PEPITA does not transmit the error signal correctly, to the point that the direction
 given by the PEPITA update rule is orthogonal to the one given by BP. In high dimensions, this can be interpreted as a random direction, meaning that PEPITA actually does not produce any meaningful weight update.
 
 However, this has to be nuanced by the fact that under a right initialization (\textit{e.g.} orthogonal initialization), PEPITA can actually transmit 
 the error signal coherently, as shown in Figure~\ref{fig:cos_sim_linear} for the orange curve.
 However it is extremely sensitive to the hyperparameters, and training was not stable for most of the 
 hyperparameters tested (as illustrated in Figure~\ref{fig:cos_sim_linear}). This is coherent with the 
 findings of \citep{srinivasan2023forward} that PEPITA can actually work on 4 to 5 hidden layers networks for some specific hyperparameters.
 Even in this case, the cosine similarity of PEPITA still decreases in the deeper layers, indicating
 a progressive perturbation of the error signal during the modulated forward pass.

 \subsubsection{Non-linear case} 
 
We extend our non‐linear training‐dynamics analysis in Figure~\ref{fig:cos_sim_nonlinear}, which plots the layer‐wise cosine similarity between FOTON’s update directions and the true backprop gradients on a 10‐layer ReLU network. We find that:
\begin{itemize}
    \item Across all ten layers, FOTON’s cosine similarity remains strictly above 0.1, even in the deepest layers—indicating that it consistently transmits a meaningful descent direction through the non‐linear activations. This is in striking contrast to PEPITA, whose cosine similarity stays near zero (and eventually becomes negative), effectively pointing its updates in random or adversarial directions.

    \item After the third hidden layer, FOTON’s cosine similarity curve flattens out rather than decaying further with depth. We conjecture that ReLU’s piecewise‐linear structure, combined with our orthogonalization, prevents additional signal erosion once the error has traversed a few layers.

    \item Because FOTON’s cosine similarity stays strictly positive and our learning rates are kept sufficiently small, Zoutendijk’s theorem \citep{nocedal1999numerical} guarantees that the algorithm will converge to a (local) minimum, which we indeed observe experimentally.

    \item To further increase this alignment, one could incorporate local Jacobian or activation‐aware corrections into the construction of $F$, as well as explore lightweight regularizers that preserve orthogonality without full Björck projections. These directions could raise the cosine similarity even closer to 1 in the non‐linear regime.
\end{itemize}
Overall, this analysis confirms that FOTON’s forward‐only updates remain a valid descent direction in deep non‐linear networks—whereas PEPITA’s updates not only lose coherence but can actively reverse the gradient—thus explaining FOTON’s ability to scale to arbitrary depths.

\begin{figure*}[t]
  \centering
  \begin{subfigure}[t]{0.48\textwidth}
    \centering
    \includegraphics[width=\textwidth]{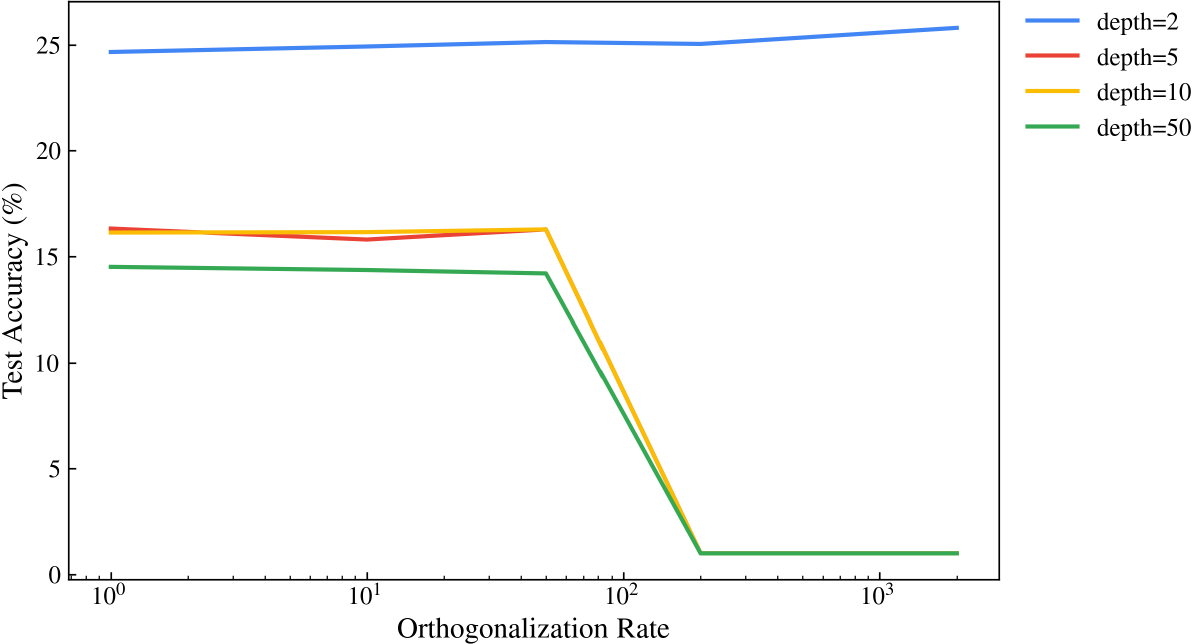}
    \caption{Test accuracy vs.\ orthogonalization rate frequency.  A rate of 200 represents roughly an orthogonalization every epoch.\\}
    \label{fig:ortho_rate}
  \end{subfigure}
  \hfill
  \begin{subfigure}[t]{0.48\textwidth}
    \centering
    \includegraphics[width=\textwidth]{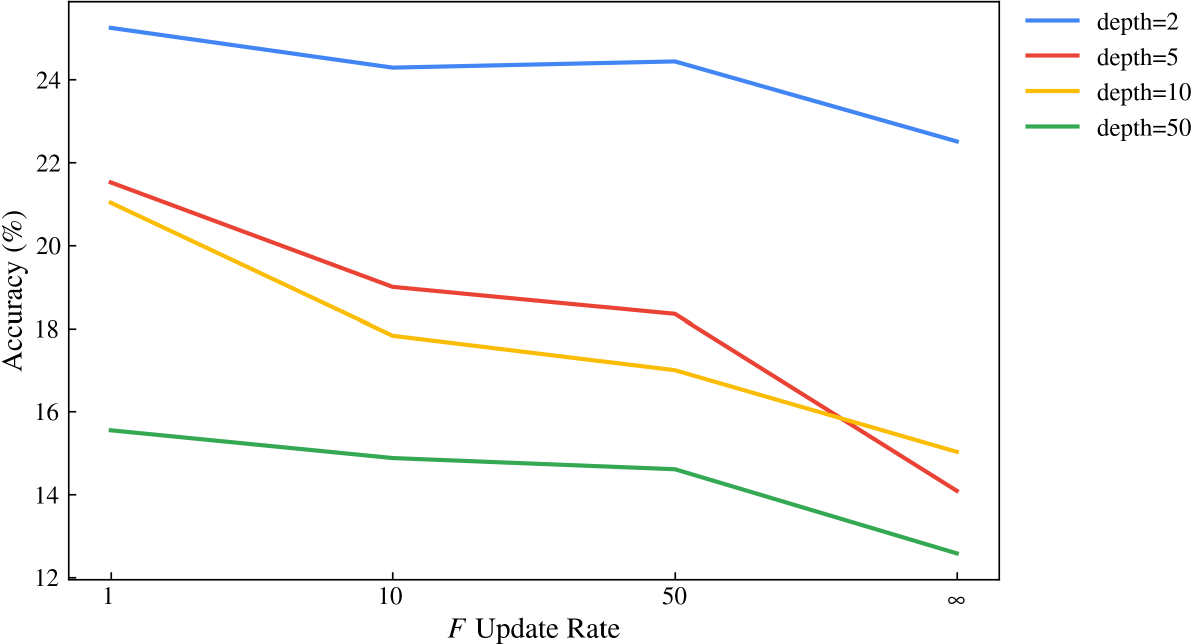}
    \caption{Test accuracy vs.\ \(F\)-update frequency.  Tick “\(\infty\)” means \(F\) was never updated after initialization.}
    \label{fig:updateF}
  \end{subfigure}
  \caption{Performance on CIFAR-100 of feedforward networks with varying depth as a function of (a) the frequency of orthogonalization and (b) the frequency of $F$-matrix updates.}
  \label{fig:ablation}
\end{figure*}
\subsection{Ablation Studies}
\label{sec:ablation}

We further explore the possibility of relaxing FOTON's theoretical constraints that we introduced in Section~\ref{sec:foton}, namely the perfect weight alignment and the exact left orthogonality of the weight matrices. To do so, we conduct ablation studies on the orthogonalization rate and its effect on network performance, and on the role of computing the exact $F$ matrix at each step. Experiments on feedforward networks of varying depth on CIFAR-100 are presented in Figure~\ref{fig:ablation}.

\subsubsection{Impact of the Orthogonalization Rate}
The results presented in Figure~\ref{fig:cos_sim_linear} show that the orthogonality constraint is necessary but its enforcement might not be needed at each update step as in the linear case PEPITA \emph{can} converge even in very deep networks, provided an orthogonal initialization. 

The appropriate rate depends on the depth and complexity of the network, with different rates potentially enabling better convergence or hampering performance. A looser orthogonality condition can be found through careful tuning, as excessively loose enforcement of orthogonality can lead to network divergence.

The data in Figure~\ref{fig:ortho_rate} reveal a clear depth‐dependent tolerance to infrequent orthogonalization. In the very shallow 2-layer model, pushing the orthogonalization interval out to 2000 steps actually yields the highest accuracy ($\approx 25.8 \%$), indicating that once weights are well-initialized, repeated re-orthogonalization can even slightly hamper learning in simple architectures. This is likely because strictly orthogonal constraints reduce the expressivity of shallow networks. This also shed light on why PEPITA might perform slightly better in the very shallow regime as seen on CIFAR-100 in Table~\ref{tab:CIFAR}.

By contrast, at depths 5, 10 and 50 the picture is very different: allowing an epoch to occur between re-orthogonalizations causes catastrophic collapse (accuracy drops to 1\%  on CIFAR-100 for intervals of 200 or 2000). This shows the paramount importance of the orthogonality constraint we propose to add to the standard PEPITA framework, without which relatively shallow networks cannot be trained.

Thus, while very shallow MLPs can “get by” on a single initial orthogonalization, deeper networks require fairly frequent enforcement (every 50 steps or less) to preserve the gradient‐norm‐preserving properties that underpin FOTON’s stability. This suggests a simple rule of thumb: increase the orthogonalization frequency with network depth. While shallow nets tolerate (and may even benefit from) sparse updates, deep nets demand regular correction to avoid divergence.

\subsubsection{Impact of the Exact Weight Alignment}

We next investigate how the sparsity of feedback-matrix updates impacts final test accuracy across different depths. Figure~\ref{fig:updateF} shows, for each depth \(d\in\{2,5,10,50\}\), the accuracy when we recompute the error-projection matrix \(F\) every \(k\) steps (\(k\in\{1,10,50,\infty\}\), where \(\infty\) denotes “no update”).

At all depths, recomputing \(F\) on \emph{every} forward pass (\(k=1\)) yields the highest accuracy, and performance drops monotonically as updates become sparser (\(k=10\), then 50, then never).  Surprisingly, even without any \(F\) refresh (\(k=\infty\)), the network still achieves non-trivial accuracy (e.g.\ 22.5\% at depth 2 and 12.6\% at depth 50).  This suggests that a single, well-initialized orthogonal feedback matrix already provides enough signal to bootstrap learning—contrasting sharply with standard PEPITA, which fails to scale beyond 5 layers.

We find no “sweet spot” at which infrequent updates match the performance of dense updates: \emph{every} configuration benefits from more frequent \(F\) refreshes.  That said, these results motivate exploring approximate or smoothed update schemes—such as weight-mirroring \citep{akrout2019deep}—to strike a better trade-off between computational cost and alignment quality, which we leave for future work.

\section{Limitations and Discussion}

Our experiments demonstrate that FOTON overcomes PEPITA's depth barrier and is a promising step towards the training of deep networks with forward-only algorithms.
Under our orthogonality constraint, FOTON transmits far more faithful learning signals than PEPITA, matching exact backprop in the linear regime and maintaining strictly positive gradient alignment even through tens of non‐linear layers. 

We also show that despite the orthogonal constraint, FOTON is able to learn better than PEPITA, even for shallow networks, and manages
to be competitive with BP on the tasks tested, without ever needing to backpropagate the error signal or to build and store an automatic-differentiation graph, dramatically cutting memory usage.

Crucially, our ablation on the $F$-matrix refresh rate shows that while recomputing $F$ every step yields the highest accuracy, infrequent updates still produce nontrivial performance. This suggests avenues for smoother or occasional updates, removing computation overhead without collapsing training.

We also show that enforcing perfect orthogonality does restrict expressivity especially in shallow models. However, we find that deeper networks benefit from the norm‐preserving and stability properties it provides. We also show that these ideas readily extend to convolutional layers via BCOP-orthogonalization, opening the door to forward‐only training of modern architectures, with a clear avenue being scaling to ResNets and Transformers.
While our empirical results support stable gradient transmission, a formal non-linear theory of gradient alignment also remains a clear direction for future work.

Finally, there is ample room to reduce overhead and further close the gap with BP, for instance by integrating adaptive optimizers, lightweight orthogonality regularizers, or approximations of $F$ updates (e.g., weight mirroring), as well as incorporating batch normalization and data augmentation. At the same time, certain standard operations such as max-pooling will require specific forward-only adaptations to fit within the FOTON framework. 

These directions promise to further close the gap with BP while retaining FOTON’s minimal‐memory, purely‐forward advantage. 
However, standard techniques like max-pooling would require a specific adaptation as a forward-only version is not straightforward to define.

\section{Conclusion}

The FOTON algorithm proposed in this work emerges as a promising forward-only alternative to 
conventional BP.
It successfully solves the main limitation of PEPITA, which is the scalability to deep networks, resorting
to the orthogonal constraint to provide a meaningful error signal to the network.
We showed the ability of FOTON not only to learn better than PEPITA in most shallow settings, but 
also to scale to deeper networks and to be competitive with BP on the tasks tested. We further show that FOTON can be adapted to convolutional networks,
opening up the possibility of applying it to more diverse architectures.

We provide a theoretical framework for FOTON, showing that it is equivalent to BP in the case of orthogonal linear networks
and can approximate it well in the non-linear case.
Our theoretical results and empirical findings are also analyzed through a thorough study of the training dynamics, underlying the 
key differences between FOTON and PEPITA in terms of the error signal transmitted to the network.

We also discuss the limitations of FOTON, notably the orthogonality constraint that could be a bottleneck for the expressiveness of the network
and the possibility of relaxing it in future research.
FOTON's performance on deep non-linear networks opens up avenues for its application in a broader range of settings.
The adaptation and use of standard techniques for BP to improve its performance is a promising direction for future research.

\clearpage
\section*{Acknowledgments}
This work was performed using HPC resources from GENCI-IDRIS (grants AD011015154 and
A0151014627), and received funding from the French Government via the program France 2030 ANR-23-PEIA-0008,
SHARP.
\bibliography{biblio}

\clearpage
\newpage
\onecolumn
\appendix

\section{Detailed Experimental Setup and Hyperparameters}
In this section, we provide a comprehensive overview of the experimental setup and hyperparameters used in our study, detailing the conditions under which \emph{FOTON} is evaluated. This includes the architectures, datasets, and training protocols chosen to highlight its practical performance compared to standard backpropagation and other forward-only baselines. 

Unless otherwise specified, the orthogonalization rate is set to 1 and the number of Björck iterations used to orthogonalize the weights is fixed to 5.
In the same way, unless otherwise specified, $F$ is updated once per epoch to keep the weight mirroring assumption valid.
The experiments were run on a single A100 GPU.

Table~\ref{tab:123} shows the hyperparameters used for the 1, 2, and 3 hidden layers networks, Table~\ref{tab:combined} shows the hyperparameters used for the 5, 10 and 50 hidden layers networks
and Table~\ref{tab:conv} shows the hyperparameters used for the convolutional networks.

\begin{table*}[ht!]
    \caption{1, 2, and 3 hidden layers network architectures and settings used in the experiments.}
    \centering
    \resizebox{\textwidth}{!}{
    \begin{tabular}{l|ccc|ccc|ccc}
        \toprule
        Hyperparameters & \multicolumn{3}{c|}{1 Hidden layer} & \multicolumn{3}{c|}{2 Hidden layers} & \multicolumn{3}{c}{3 Hidden layers} \\
        \midrule
           Dataset & MNIST & CIFAR10 & CIFAR100 & MNIST & CIFAR10 & CIFAR100 & MNIST & CIFAR10 & CIFAR100 \\
          \midrule
          Input size & $28 \times 28 \times 1$ & $32 \times 32 \times 3$ & $32 \times 32 \times 3$ & $28 \times 28 \times 1$ & $32 \times 32 \times 3$ & $32 \times 32 \times 3$ & $28 \times 28 \times 1$ & $32 \times 32 \times 3$ & $32 \times 32 \times 3$ \\
          Hidden units & $1 \times 1024$ & $1 \times 1024$ & $1 \times 1024$ & $2 \times 1024$ & $2 \times 1024$ & $2 \times 1024$ & $3 \times 1024$ & $3 \times 1024$ & $3 \times 1024$ \\
          Output units & 10 & 10 & 100 & 10 & 10 & 100 & 10 & 10 & 100 \\
          \midrule
          Learning rate BP & $0.1$ & $0.01$ & $0.1$ & $0.1$ & $0.01$ & $0.1$& $0.1$ & $0.01$ & $0.1$  \\
          Learning rate PEPITA & $0.1$ & $0.2$ & $0.01$ & $0.1$ & $0.01$ & $0.01$ & $0.001$ & $0.01$ & $0.01$ \\
          Learning rate FOTON & $0.2$ & $0.05$ & $0.05$ & $0.2$ & $0.05$ & $0.05$ & $0.1$ & $0.05$ & $0.05$ \\
          Weight decay BP & 0 & 0 & 0 & 0 & 0 & 0 & 0 & 0 & 0 \\
          Weight decay PEPITA & $10^{-5}$ & $10^{-4}$ & $10^{-5}$ & $10^{-5}$ & $10^{-4}$ & $10^{-5}$ & $10^{-5}$ & $10^{-4}$ & $10^{-4}$ \\
          Weight decay FOTON & $0$ & $0$ & $0$ & $10^{-2}$ & $10^{-1}$ & $0$ & $0$ & $0$ & $10^{-3}$ \\
          Batch size BP & $64$ & $64$ & $64$ & $64$ & $64$ & $64$ & $64$ & $64$ & $64$ \\
          Batch size PEPITA & $64$ & $64$ & $64$ & $64$ & $64$ & $64$ & $64$ & $64$ & $64$ \\
          Batch size FOTON & $256$ & $256$ & $256$ & $256$ & $256$ & $256$ & $256$ & $256$ & $256$ \\
          Number of Epochs & 100 & 100 & 100 & 100 & 100 & 100 & 100 & 100 & 100 \\
          Dropout BP & 0 & 0 & 0 & 0 & 0 & 0 & 0 & 0 & 0 \\
          Dropout PEPITA & $0.1$ & $0.1$ & $0.1$ & $0.1$ & $0.1$ & $0.1$ & $0.1$ & $0.1$ & $0.1$ \\
          Dropout FOTON & $0$ & $0.1$ & $0.2$ & $0$ & $0.2$ & $0.2$ & $0$ & $0$ & $0$ \\
          Error PEPITA & MSE & MSE & MSE & MSE & MSE & MSE & MSE & MSE & MSE \\
          Error FOTON & MSE & CE & CE & MSE & CE & CE & MSE & CE & CE \\
          CE temperature &   & $4$ & $1$ &   & 4 & $1$ &   & $1$ & $2$ \\
          Decay epochs BP & None & None & None & None & None & None & None & None & None \\
          Decay epochs PEPITA & $60, \ 90$ & $60, \ 90$ & $60, 90$ & $60, \ 90$ & $60, \ 90$ & $60, 90$ & $60, \ 90$ & $60, \ 90$ & $60, 90$ \\
          Decay epochs FOTON & None & None & None & None & None & None & None & None & None \\
        \bottomrule
    \end{tabular}}
    \label{tab:123}
\end{table*}

 \begin{table*}[ht!]
    \caption{5, 10 and 50 hidden layers network architectures and settings used in the experiments.}
    \centering
    \resizebox{\textwidth}{!}{
    \begin{tabular}{l|ccc|ccc|ccc}
        \toprule
         Hyperparameters &\multicolumn{3}{c|}{5 Hidden layers}& \multicolumn{3}{c|}{10 Hidden layers} & \multicolumn{3}{c}{50 Hidden layers} \\
         \midrule
          Dataset & MNIST & CIFAR10 & CIFAR100  & MNIST & CIFAR10 & CIFAR100 & MNIST & CIFAR10 & CIFAR100 \\
         \midrule
          Input size & $28 \times 28 \times 1$ & $32 \times 32 \times 3$ &  $32 \times 32 \times 3$ & $28 \times 28 \times 1$ & $32 \times 32 \times 3$ & $32 \times 32 \times 3$ & $28 \times 28 \times 1$ & $32 \times 32 \times 3$ &  $32 \times 32 \times 3$  \\
          Output units & 10 & 10 & 100  & 10 & 10 & 100 & 10 & 10 & 100 \\
         \midrule
          Learning rate BP &  $0.1$ & $0.01$ & $0.1$ & $0.1$ & $0.01$ & $0.1$ &  $0.01$ & $0.01$ & $0.01$\\
          Learning rate FOTON & $0.2$ & $0.01$ & $0.01$  & $0.01$ & $0.01$ & $0.01$& $0.1$ & $0.005$ & $0.005$  \\
          Weight decay BP & 0 & 0 & 0 & 0 & 0 & 0&$10^{-4}$&$10^{-4}$&$10^{-4}$  \\
          Weight decay FOTON & $0$ & $0$ & $10^{-4}$ & $0$ & $0$ & $10^{-4}$& $10^{-4}$ & $10^{-4}$ & $10^{-4}$  \\
          Batch size BP & $64$ & $64$ & $64$  & $64$ & $64$ & $64$ & $64$ & $64$ & $64$  \\
          Batch size FOTON & $256$ & $256$ & $256$ & $256$ & $256$ & $256$ & $256$ & $256$ & $256$ \\
          Number of Epochs & 100 & 100 & 100  & 100 & 100 & 100 & 100 & 100 & 100 \\
          Dropout BP & 0 & 0 & 0  & 0 & 0 & 0& 0 & 0 & 0   \\
          Dropout FOTON & $0$ & $0$ & $0$  & $0$ & $0$ & $0$& 0 & 0 & 0   \\
          Error FOTON & MSE & CE & CE  & MSE & CE & CE & MSE & CE & CE\\
          CE temperature &   & $1$ & $2$  &   & $2$ & $2$ &   & $2$ & $2$\\
          Decay epochs BP & 30,60 & 30,60 & 30,60  & 30,60 & 30,60 & 30,60 & 30,60 & 30,60 & 30,60 \\
          Decay epochs FOTON & 30,60 & 30,60 & 30,60& 30,60 & 30,60 & 30,60& 30,60 & 30,60 & 30,60  \\
        \bottomrule
    \end{tabular}}
    \label{tab:combined}
\end{table*}
\begin{table*}[ht!]
    \caption{$1$ and $2$ convolutional layers followed by a $1$ linear layer classifier head architectures and settings used in the experiments.}
    \centering
    \small
    \begin{tabular}{l|cc|cc}
        \toprule
         Hyperparameters & $1$ convolutional layer & & $2$ convolutional layers   \\
         \midrule
          Dataset & MNIST & CIFAR100  & MNIST  & CIFAR100  \\
         \midrule
          Input size & $28 \times 28 \times 1$ &   $32 \times 32 \times 3$ & $28 \times 28 \times 1$ &  $32 \times 32 \times 3$ \\
          Output units & 10 & 100 & 10  & 100  \\
         \midrule
          Learning rate BP &   &  & &   \\
          Learning rate FOTON & $0.08$ & $0.08$ & $0.06$  & $0.01$   \\
          Weight decay BP & 0 & 0 & 0 & 0  \\
          Weight decay FOTON & $0$ & $0$ & $0$ & $0$   \\
          Batch size BP & $64$ & $64$ & $64$  & $64$   \\
          Batch size FOTON & $256$ & $256$ & $256$ & $256$   \\
          Number of Epochs & 100 & 100 & 100  & 100   \\
          Dropout BP & 0 & 0 & 0  & 0   \\
          Dropout FOTON & $0$ & $0$ & $0$  & $0$   \\
          Error FOTON & CE & CE & CE  & CE  \\
          CE temperature & $1$  & $1$ & $2$  &  $1$  \\
          Decay epochs BP & None & None & None  & None   \\
          Decay epochs FOTON & None & None & None  & None   \\
        \bottomrule
    \end{tabular}
    \label{tab:conv}
\end{table*}
In the first case reported Table~\ref{tab:123}, we used a hidden size of 1024 hidden units, 
either for 1, 2, or 3 hidden layers, and the output size was 10 for MNIST and CIFAR10, and 100 for CIFAR100.
In the second case reported in Table~\ref{tab:combined}, we used a hidden size of 
256 units, and the output size was 10 for MNIST and CIFAR10, and 100 for CIFAR100.
In the third case reported in Table~\ref{tab:conv}, we used one or two convolutional layer with 32 filters of size $3 \times 3$ followed by a linear layer, after each convolution we applied a rescaled avarage pooling and the output size was 10 for MNIST and 100 for CIFAR100.
The BP baseline was trained using standard SGD optimizer. The ReLU non linearity was used between every layer in every setting. However, for the deeper architecture with 50 layers we replaced ReLU with TanH, since preliminary trials showed that TanH helped mitigate the dying ReLU effect and consistently provided better results in terms of both convergence and final accuracy.

\section{Theoretical comparison between FOTON and Backpropagation}
In this section we prove in details that FOTON gives a first order approximation of the exact gradient computed by Backpropagation. 

\subsection{Equivalence in deep linear regime} \label{sec:linear_proof}

We start showing the simplified result in the case of deep linear networks that are neural networks without activation functions. These models are just a composition of linear functions and therefore they do not have any approximation advantage over a linear regression, however they differ in the optimization dynamics and while PEPITA, and its variants, match backpropagation in the linear regression case they differ from exact gradient computation on linear deep neural networks. We show that FOTON computes exactly the gradients in this case and experiments support our theoretical findings. We believe that the simplified case will make more clear the role of orthogonality and of FOTON's learning rule.

\begin{theorem}[Equivalence in deep linear regime]\label{th: FOTON_lin}
    In the case of deep linear neural networks, FOTON perfectly matches backpropagation, i.e.
    \begin{align*}
        \delta W_\ell^{FO} = \delta W_\ell^{BP} = \frac{\partial \mathcal{L}}{\partial W_\ell} \quad \forall \ell = 1, \ldots, L
    \end{align*}
\end{theorem}
\begin{proof}
    We start recalling the update of backpropagation:
    \begin{align}
        \delta W_\ell ^{BP} & = - \eta \: \delta a_{\ell} ^{BP} (h_{\ell-1})^\top \;
        \text{ for } 1\leq \ell< L; \\
        \delta W_L ^{BP} & = - \eta \: e \: h_{L-1}^\top \: . \label{eq:bp}
    \end{align}

    It follows that FOTON is equivalent to backpropagation
    if and only if $\delta a_\ell ^{FO} = \delta a_{\ell}^{BP}$ 
    for all $\ell$. 
    The two recursive formulas can be expanded as follows:
    \begin{align*}
        \delta a_\ell ^{FO}=& W_\ell\sigma_{\ell-1}(W_{\ell-1}\sigma_{\ell-2}( \cdots \sigma_1(W_1x))) - W_\ell\sigma_{\ell-1}(W_{\ell-1}\sigma_{\ell-2}( \cdots \sigma_1(W_1(x-Fe))))
        \\ \delta a_\ell ^{BP}=&  \sigma'_\ell W_{\ell + 1}^\top \sigma'_{\ell+1}\cdots W_L^\top e
    \end{align*}

    In the linear case, that is when the activation function are the identity, the two become:
    \begin{align*}
        \delta a_\ell ^{FO}=& W_\ell(W_{\ell-1}( \cdots W_1x)) - W_\ell(W_{\ell-1}( \cdots W_1(x-Fe)))
        \\ \delta a_\ell ^{BP}=&  W_{\ell + 1}^\top W_{\ell + 2}^\top \cdots \cdots W_L^\top e
    \end{align*}
    When the weight matrices are row-orthogonal, the first update of FOTON is the following:
    \begin{align*}
        \delta a_\ell ^{FO}=& W_\ell(W_{\ell-1}( \cdots W_1Fe)) 
    \end{align*}
    and admitting a perfect weight alignment, as implemented in FOTON, since 
    $F = W_1^\top W_2^\top \cdots W_L^\top$, we have:
    \begin{align*}
        \delta a_\ell ^{FO}=& W_\ell(W_{\ell-1}( \cdots W_1 W_1^\top W_2^\top \cdots W_L^\top)e)
        = W_{\ell + 1}^\top W_{\ell + 2}^\top \cdots \cdots W_L^\top e = \delta a_\ell ^{BP}
    \end{align*}
    and the proof is complete.
\end{proof}

\subsection{Equivalence in deep norm-preserving regime}
FOTON may perfectly match backpropagation in a more
general and interesting case than the previously discussed linear one but at the cost
of a bigger memory consumption. This will not make FOTON a valuable alternative
to the better performing backpropagation but it shows a closer gap to exact gradient computation with respect to PEPITA.

\begin{theorem}\label{th: FOTON_piecewise}    
    Let $f$ be an orthogonal neural network with piecewise orthogonal (and linear) activation functions,
    such as GroupSort. We assume that the non-linear activations are incorporated in the matrix $F$ as
    \begin{align*}
        F = W_1^\top \sigma_1' W_2^\top \sigma_2' \cdots W_L^\top
    \end{align*} 
    where the gradient of the activation functions are taken in the respective activation $h_\ell$ obtained
    from the standard forward pass of $x$. 
    In this case FOTON is exactly equivalent to backpropagation, i.e.
    \begin{align*}
        \delta W_\ell^{FO} = \delta W_\ell^{BP} \quad \forall \ell
    \end{align*}
\end{theorem}
\begin{proof}
    The proof is similar to the one given in \ref{th: FOTON_lin} with some additional details to take
    care of. First of all, the error $e$ can be considered small compared to $x$, i.e. 
    $\lVert Fe \rVert \ll \lVert x \rVert$. We can thus assume that the activation
    patterns by $x$ do not change from those by $x-Fe$ and so we have that 
    \begin{align*}
        \delta a_\ell ^{FO} &= W_\ell\sigma_{\ell-1}(W_{\ell-1}\sigma_{\ell-2}( \cdots \sigma_1(W_1x))) - W_\ell\sigma_{\ell-1}(W_{\ell-1}\sigma_{\ell-2}( \cdots \sigma_1(W_1(x-Fe))))
        \\ &= W_\ell\sigma_{\ell-1}(W_{\ell-1}\sigma_{\ell-2}( \cdots \sigma_1(W_1(Fe))))
    \end{align*}
    and thus we have 
    \begin{align*}
        \delta a_\ell ^{FO} &= W_\ell\sigma_{\ell - 1}(W_{\ell-1}\sigma_{\ell-2}( \cdots \sigma_1 W_1 W_1^\top \sigma_1 'W_2^\top \sigma_2'\cdots \sigma_{L-1}' W_L^\top)e) \\
        &= \sigma_{\ell}' W_{\ell + 1}^\top \sigma_{\ell + 1}' \cdots \cdots \sigma_{L-1}' W_L^\top e \\
        &= \delta a_\ell ^{BP}
    \end{align*}
    since the activation $\sigma_{\ell}$ are piecewise orthogonal $\forall \ell$.
\end{proof}

\begin{remark}
    The main issue with the previous theorem is that the gradients w.r.t the activations depend on the single data points. Therefore in order to implement this version, we need to store a matrix F for
    any data point (or equivalently a tensor of size the dimension of F times the batch size) obtaining an algorithm
    with a memory cost comparable to backpropagation. A main advantage of FOTON is to have a memory cost
    independent of the batch size in the error projection, which is a major advantage in the case of large batch sizes.
\end{remark}

\subsection{Linear approximation in deep non-linear regime}
\label{sec:nonlinear_proof}

In the case of non-linear networks, there is no guarantee for FOTON to match backpropagation. However some approximations allow us to see how it can still approach backpropagation and make the model learn.

Let $f$ be an orthogonal neural network with any non-linear activation functions $\sigma_\ell$. If we approximate the activation functions linearly ($\sigma_\ell(x) \approx \lambda x$), we get:
\begin{align*}
    \delta h_\ell ^{FO} &= \sigma_l(W_\ell\sigma_{\ell-1}(W_{\ell-1}\sigma_{\ell-2}( \cdots \sigma_1(W_1x)))) - \sigma_l(W_\ell\sigma_{\ell-1}(W_{\ell-1}\sigma_{\ell-2}( \cdots \sigma_1(W_1(x-Fe))))) \\
    &\approx \lambda W_\ell \lambda W_{\ell-1} \lambda \cdots \lambda W_1 F e \\
    &= \lambda^{\ell} \cdot W_{\ell+1}^T W_{\ell+2}^T \dots W_L^T \nabla_L \mathcal{L}
\end{align*}
and
\begin{align*}
    \delta h_\ell ^{BP} &=  W_{\ell + 1}^\top \sigma'_{\ell+1}\cdots W_L^\top e 
    \approx \lambda^{L-\ell-1} \cdot W_{\ell+1}^T W_{\ell+2}^T \dots W_L^T \nabla_L \mathcal{L}
\end{align*}

This leads to:
\begin{align*}
    \delta h_\ell ^{FO} &\approx \lambda^{2\ell-L+1} \cdot \delta h_\ell ^{BP}
\end{align*}
This indicates that the FOTON learning rule can still provide a descent direction in the non-linear case. This is further confirmed in our experiments, where FOTON performs close to backpropagation in many different settings.

\section{Extension of FOTON to Convolutional Networks}
\label{sec:convolution_ext}

As mentioned in the main text, the application of FOTON to convolutional networks
is non-trivial, as the orthogonality constraint is applied to the Toeplitz matrix and not to the convolutional
kernel.
In the original PEPITA paper~\citep{dellaferrera2022error}, a seminal proposition to update convolutional layers relying on forward-only algorithm was made.
The kernels are learned based on post and pre-synaptic related terms and the update is computed for each filter.
When writing the convolutional layer as a linear operation (the corresponding matrix is a concatenation of doubly Toeplitz matrices as shown by \citet{2020arXiv200607117Y}), this update does not match the original PEPITA update, as the update is not computed on the kernel indices but on the kernel patches.
In FOTON, we propose to apply the same update as in the fully connected layers.
We designed the convolutional update of FOTON to match backpropagation in the linear (or piecewise linear) case as in section \ref{sec:linear_proof}.
In this section, we first recall the update rule for backpropagation and derive the update rule for FOTON,
using the analogy built in section \ref{sec:linear_proof}.

For an MLP, BP updates the parameters as:
$
    \delta W_\ell = \frac{\partial \mathcal{L}}{\partial W_\ell} = \frac{\partial \mathcal{L}(W_\ell h_{l-1})}{\partial W_\ell} = \delta a_\ell h_{l-1}^T
$,
where $\mathcal{L}$ is the loss function. This follows from the fact that
\begin{align*}
    \partial \mathcal{L}((W_\ell + \mathcal{E}) h_{l-1}) &= \partial \mathcal{L}(W_\ell h_{l-1}) + \left\langle \frac{\partial \mathcal{L}}{\partial W_\ell h_{l-1}}, \mathcal{E} h_{l-1} \right\rangle + o(\lVert \mathcal{E} \rVert) \\
    &= \partial \mathcal{L}(W_\ell h_{l-1}) + \left\langle \frac{\partial \mathcal{L}}{\partial W_\ell h_{l-1}} h_{l-1}^\top, \mathcal{E} \right\rangle + o(\lVert \mathcal{E} \rVert)
\end{align*}
and the result follows from the definition of differential.

In the case of convolutional layers, the $W_\ell h_{\ell-1}$ is replaced
by the convolution $k_\ell \ast h_{\ell-1}$ where $k_\ell$ is the kernel.
It follows that: 
\begin{align*}
     \mathcal{L}((k_\ell + \mathcal{E}) \ast h_{l-1}) &=  \mathcal{L}(k_\ell \ast h_{l-1}) + \left\langle \frac{\partial \mathcal{L}}{\partial k_\ell \ast h_{l-1}}, \mathcal{E} \ast h_{l-1} \right\rangle + o(\lVert \mathcal{E} \rVert) \\
    &=  \mathcal{L}(W_\ell \ast h_{l-1}) + \left\langle \frac{\partial \mathcal{L}}{\partial W_\ell \ast h_{l-1}}, h_{l-1} \ast \mathcal{E} \right\rangle + o(\lVert \mathcal{E} \rVert)
    \\ &=  \mathcal{L}(W_\ell \ast h_{l-1}) + \left\langle \frac{\partial \mathcal{L}}{\partial W_\ell \ast h_{l-1}}, \text{conv}_{h_{l-1}}(\mathcal{E}) \right\rangle + o(\lVert \mathcal{E} \rVert)
    \\ &=   \mathcal{L}(W_\ell \ast h_{l-1}) + \left\langle \text{conv}_{h_{l-1}}^{\star} \left( \frac{ \partial \mathcal{L}}{\partial W_\ell \ast h_{l-1}} \right) , \mathcal{E} \right\rangle + o(\lVert \mathcal{E} \rVert)
\end{align*}
where $\text{conv}_k(x) = k \ast x$ is the left-convolution by a fixed kernel $k$. Since it's a linear
operator, it admits an adjoint, the \textit{transposed convolution}, that we denote by $\star$. It follow that the update rule for the kernel in backpropagation is:
\begin{align*}
    \delta k_\ell =  \text{conv}_{h_{\ell-1}}^{\star}(\delta a_\ell)
\end{align*}
To match BP in the linear and piecewise linear case, FOTON updates the
kernel as follows:
\begin{align*}
    \delta k_\ell &=  \text{conv}_{h_{\ell-1}}^{\star}(\delta a_\ell^{FO}) 
    \\ &=  \text{conv}_{h_{\ell-1}}^{\star}(a_\ell - a_\ell^{err}) 
\end{align*}

In practice, the $\text{conv}_{h_{\ell}}^{\star}$ operator corresponds to the \texttt{torch.nn.grad.conv2d\_input} function in PyTorch.

\section{Choice of Pooling Module}
Pooling modules are important parts of convolutional neural networks and they are
often considered as non linear activations so the theory developed in \ref{sec:nonlinear_proof}
applies identically. Max pooling and average pooling are among the most used pooling modules.
Adapting max pooling to any forward-only learning system, is however not straightforward. 
The nature of max pooling itself, which selects the maximum value from a set of inputs and discards the rest is inherently non-differentiable.
Handling its selective nature and data-dependency cannot be done only by keeping track of the maximum value as is done in backpropagation.

The alternative we used is average pooling, which is differentiable more compatible with FOTON's forward-only error-driven modulation. 
Average pooling is linear and row orthogonal if scaled by $\sqrt{2}$, which we implemented to keep the FOTON's orthogonality constraint.

However, it is important to note that in practice, max pooling tends to outperform average pooling in terms of performance
in most computer vision tasks, particularly in image classification tasks with datasets such as CIFAR10 and CIFAR100.
Adapting max-pooling to FOTON could be a potential future research direction, as it would align better with the standard practice in the field.

\end{document}